\documentclass[10pt,twocolumn,letterpaper]{article}
\usepackage[utf8]{inputenc} 
\usepackage[T1]{fontenc}   
\usepackage{iccv}              

\definecolor{iccvblue}{rgb}{0.21,0.49,0.74}
\usepackage[pagebackref,breaklinks,colorlinks,allcolors=iccvblue]{hyperref}

\usepackage{caption}
\usepackage{subcaption}
\usepackage{graphicx} 
\usepackage[american]{babel}
\usepackage{microtype}
\usepackage{multirow}
\usepackage{booktabs}
\usepackage{algorithm}

\usepackage{algorithm}
\usepackage{algorithmicx}
\usepackage{algpseudocode}

\usepackage{bm}

\usepackage{wrapfig}
\usepackage{pifont}
\usepackage{multirow}
\usepackage[export]{adjustbox}
\usepackage{color}
\usepackage{colortbl}
\usepackage{lipsum}
\usepackage{pifont}       
\usepackage{bbding}    
\usepackage[dvipsnames]{xcolor}
\usepackage{array}

\usepackage[capitalize]{cleveref}
\usepackage{float}
\crefname{section}{Sec.}{Secs.}
\Crefname{section}{Section}{Sections}
\Crefname{table}{Table}{Tables}
\crefname{table}{Tab.}{Tabs.}
\definecolor{red}{RGB}{255,0,0}
\definecolor{blue}{RGB}{0,0,255}
\definecolor{green}{RGB}{0,255,0}
\definecolor{mygray}{gray}{.9}
\definecolor{mygray2}{gray}{.5}
\definecolor{mywarning}{RGB}{233,144,61}
\definecolor{mygreen}{RGB}{93,174,86}
\definecolor{codefunc}{RGB}{73,122,234}

\definecolor{mygreen}{RGB}{0,154,85}
\definecolor{myy}{RGB}{126,95,0}
\definecolor{myred}{RGB}{212,121,116}
\definecolor{myblue}{RGB}{184, 134, 73}
\definecolor{mynewgreen}{RGB}{113,188,169}
\definecolor{mypurple}{RGB}{123,104,238}
\colorlet{R1}{myblue}
\colorlet{R2}{mypurple}
\colorlet{R3}{myred}
\colorlet{R6}{mypurple}
\definecolor{mycite}{RGB}{73,123,184}
\colorlet{cite}{mycite}

\newcommand{\x}{{\bm{x}}}
\newcommand{\y}{{\bm{y}}}
\newcommand{\X}{{\bm{X}}}
\newcommand{\Y}{{\bm{Y}}}
\newcommand{\gt}{{\bm{T}}}
\newcommand{\h}{{\bm{h}}}
\newcommand{\D}{{\textrm{d}}}

\newcommand{\oA}{\bar A}
\newcommand{\ob}{\bar\B}
\newcommand{\B}{{\bm{b}}}
\newcommand{\C}{{\bm{c}}}

\newcommand{\T}{{\top}}

\def\paperID{5263} 
\def\confName{ICCV}
\def\confYear{2025}

\title{BS-Mamba for Black-Soil Area Detection On the Qinghai-Tibetan Plateau}

\author{Xuan Ma$^1$, Zewen Lv$^1$, Chengcai Ma$^1$, Tao Zhang$^2$, Yuelan Xin$^{3}$, and Kun Zhan$^{1,4,\star}$\\
	1. School of Information Science and Engineering, Lanzhou University\\
	2. State Key Laboratory of HIGAE, College of Ecology, Lanzhou University\\
	3. School of Physics and Electronic Information Engineering, Qinghai Normal University\\
	4. Key Laboratory of AI and Information Processing, Hechi University\\
	{\small \url{https://github.com/kunzhan/BS-Mamba}}}

\begin{document}
\maketitle
\begin{abstract}
	Extremely degraded grassland on the Qinghai-Tibetan Plateau (QTP) presents a significant environmental challenge due to overgrazing, climate change, and rodent activity, which degrade vegetation cover and soil quality. These extremely degraded grassland on QTP, commonly referred to as black-soil area, require accurate assessment to guide effective restoration efforts. In this paper, we present a newly created QTP black-soil dataset, annotated under expert guidance. We introduce a novel neural network model, BS-Mamba, specifically designed for the black-soil area detection using UAV remote sensing imagery. The BS-Mamba model demonstrates higher accuracy in identifying black-soil area across two independent test datasets than the state-of-the-art models. This research contributes to grassland restoration by providing an efficient method for assessing the extent of black-soil area on the QTP.
\end{abstract}
\section{Introduction}
The Qinghai-Tibetan Plateau (QTP) is an ecological region that significantly influences climate regulation and water security in Asia, while also global carbon cycling and serves as a sanctuary for numerous unique species~\cite{CAO2019988}. However, QTP has faced increasing concern due to extremely grassland degradation.
Restoring these degraded grassland, such as the black-soil-type areas on the QTP, is a significant challenge, often requiring over two centuries to restore ecosystem health~\cite{daily1995restoring}.

Black-soil patches, characterized by the absence of mattic epipedon~\cite{shang2008effect}, are recognized as a unique ecological phenomenon that primarily occurs within the grasslands of QTP~\cite{SHANG2018}.
The distribution of black-soil area on the QTP is expanding, primarily concentrated in the cold and extreme environments at altitudes ranging between 3,000 and 5,000 meters~\cite{SHANG2018}.
Black-soil area detection inherently involves identifying the extent of mattic epipedon, as the two conditions are mutually exclusive and collectively exhaustive. Given this binary nature of the relationship between black-soil area and mattic epipedon, we have developed a QTP black-soil dataset using the unmanned aerial vehicle (UAV) imagery, carefully annotating black-soil patches and mattic epipedon under the guidance of ecological experts.
Exploiting this dataset, we train a model to infer black-soil area, facilitating the evaluation of restoration strategies and monitoring grassland quality on the QTP.

To enhance detection capabilities, we aim to design a binary segmentation neural network specifically for black-soil area detection.
U-Net~\cite{ronneberger2015u,huang2020unet,zhang2024vegetation} is widely recognized for its outstanding performance in binary segmentation tasks. While \textit{vanilla} U-Net~\cite{ronneberger2015u} excels at extracting local pixel-wise features through convolutional operations, it has limitations in capturing long-range dependencies between correlated pixels within the same class.
In contrast, Mamba~\cite{gu2023mamba}, which utilizes patch-based sequence processing, are particularly effective at capturing long-range dependencies between different patches.
Recent advancements in vision Mamba~\cite{zhang2024survey,xu2024survey,zhang2024vm} have shown superior performance over both convolutional neural networks and vision transformers~\cite{dosovitskiy2020image} in many computer vision tasks. Mamba, known for its efficiency in handling long sequences and capturing global contextual information as a state-space model (SSM)~\cite{gu2023mamba}, has emerged as a highly effective alternative to transformers. Thus, vision Mamba models have achieved state-of-the-art performance in various computer vision tasks. Building on recent innovations of vision Mamba~\cite{zhang2024survey,xu2024survey,zhang2024vm} and hybrid structures, we introduce BS-Mamba, a novel black-soil area detection network that leverages the strengths of the Mamba~\cite{gu2023mamba} and incorporates skip connections from multi-branch encoder features to the decoder, enhancing detection performance overall.

In this paper, we propose a BS-Mamba model, which integrates a multi-branch parallel feature fusion module into Mamba.
This model combines low-level and high-level features by injecting semantic information into the low-level features and refining the high-level ones with detailed spatial information.
The multi-branch layer-wise parallel structure compensates for the loss of feature resolution typically seen in Mamba.
Inspired by the multi-branch layer-wise parallel design, the multi-scale features extracted by Mamba are upsampled and combined with the different-resolution features from the encoder, allowing for precise exploring long-range dependencies.

\begin{figure}[!t]
	\centering
	\includegraphics[width=0.96\linewidth]{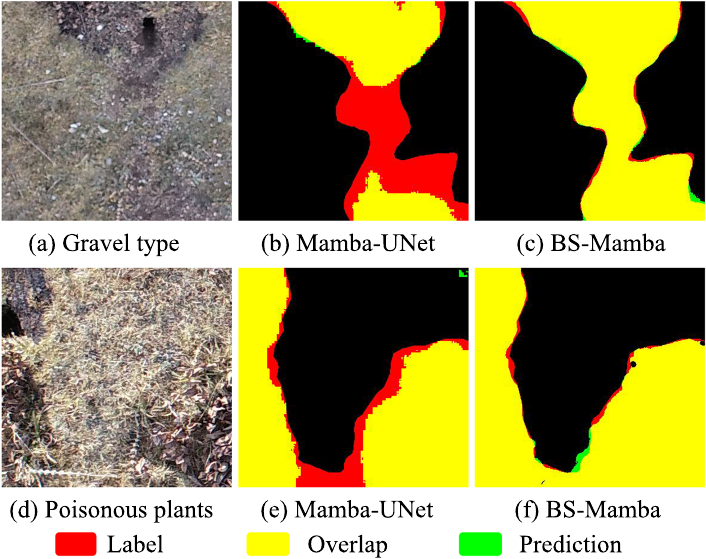}
	\caption{Comparison of our BS-Mamba model and a variant of Mamba, Mamba-Unet, in detecting black-soil areas. In both gravel-type and poisonous-plant-type regions, our BS-Mamba model consistently outperforms Mamba-Unet. Detection challenges include omissions, inaccuracies, and unclear boundaries. These issues arise mainly due to the similar color tones between black-soil patches and mattic epipedon (first row) and the high diversity of invasive weeds (second row). The results demonstrate that our BS-Mamba model effectively mitigates these challenges.
	}\label{fig:challenges}
\end{figure}
Figure~\ref{fig:challenges} illustrates the black-soil area detection results of our BS-Mamba model compared to a Mamba variant, Mamba-UNet\cite{wang2024mamba}. Detection with Mamba variants faces challenges due to several factors: (1) the varying stages of degradation introduce significant diversity among black-soil types, complicating accurate identification; (2) the natural boundaries between grasslands and black-soil patches are often unclear. For instance, gravel-type black-soil patches~\cite{XBNX202302004} seamlessly blend with surrounding grassy areas, as shown in the first row of Figure~\ref{fig:challenges}, leading to misclassifications and omissions. Additionally, dense weed cover, the presence of poisonous plants\cite{shang1157}, or the smooth surfaces of weed-covered areas further impair detection accuracy, as depicted in the second row of Figure~\ref{fig:challenges}. In contrast, the third column of Figure~\ref{fig:challenges} demonstrates that our BS-Mamba model effectively addresses these challenges, achieving significant improvements in accuracy and boundary clarity.

BS-Mamba retains the strengths of Mamba while enhancing the segmentation of the black-soil patches and mattic epipedon. Empirical results show that our hybrid architecture, which integrates convolutional blocks with Mamba blocks, outperforms methods based solely on standalone convolutional blocks or Mamba blocks. This hybrid approach effectively combines the high-resolution spatial features from convolutional blocks with the global context captured by Mamba. Unlike Mamba-UNet, which only uses Mamba blocks within a U-shaped network architecture, our BS-Mamba model fuses both convolutional and Mamba features within a two-branch encoder framework. This design enables more accurate and efficient detection of black-soil areas.
We conduct extensive experiments on the proposed QTP-BS dataset, showing the outstanding performance of the BS-Mamba model in precise black-soil patches and mattic epipedon segmentation tasks.
To further validate the model's generalization ability, we also perform comprehensive testing on a secondary test set.
The results conclusively demonstrate the strong adaptability and stability of the BS-Mamba model under varying environmental conditions.

The main contributions are summarized as follows:

(1) We have assembled a large-scale dataset of black-soil grasslands on the QTP, named the QTP-BS dataset. To the best of our knowledge, this is the first dataset  created specifically for the challenging task of black-soil area detection. The QTP-BS dataset includes 15,000 images of classic black-soil areas in the QTP grasslands, along with their corresponding ground-truth labels.

(2) We proposed a BS-Mamba model, which innovatively combines Mamba and convolution, leveraging the high-resolution spatial details of convolution features and the long-range dependencies encoded by Mamba. Experiments demonstrate that BS-Mamba performs exceptionally well in black-soil patches and mattic epipedon detection.

(3) Extensive experiments demonstrate the outstanding performance of our proposed BS-Mamba model on the challenging QTP-BS dataset. The model achieves state-of-the-art results. These results underscore its robustness and ability to generalize effectively across diverse and complex ecological scenarios.

The rest of this paper is organized as follows: \S\ref{sec:bs-mamba} introduces the fundamental principles of the state-space model and explains how BS-Mamba can be applied to black-soil area detection. \S\ref{sec:data} provides a detailed description of the QTP-BS dataset. \S\ref{sec:exp} presents the experimental results, comparing the model performance with current state-of-the-art methods. Finally, \S\ref{sec:con} concludes the paper.
\section{BS-Mamba for black-soil area detection}\label{sec:bs-mamba}
\subsection{Task Setup and Denotation}
Our goal is to assess the impact of black-soil area degradation on the QTP. To achieve this, we develop a black-soil dataset and trained a Mamba-based model, BS-Mamba, to infer black-soil pixels accurately. The assembled black-soil dataset consists of high-resolution images captured through UAV remote sensing. These  high-resolution images are uniformly divided into smaller size $H \times W$, creating a training dataset $\mathcal D = \{(\X_1,\gt_1), (\X_2,\gt_2), \ldots, (\X_n,\gt_n)\}$ where $\X\in\mathbb{R}^{H\times W\times3}$ is an image with three RGB channels and $\gt\in\mathbb R^{H \times W\times2}$ is the ground-truth label.
$\X$ is input into the black-soil patches detection network BS-Mamba $\Y=\sigma(\X|\theta)$ and it generates an output prediction $\Y\in\mathbb R^{H\times W\times2}$.
where $\sigma(\cdot|\theta)$ denotes the BS-Mamba neural network and $\theta$ is its parameter. $\gt$ and $\Y$ have two channels: one for black-soil patches and the other for mattic epipedon.
\subsection{State-Space Model}
A linear dynamic system are used to map an input signal $x(t) \in \mathbb{R}$ to an output $y(t) \in \mathbb{R}$ through  latent states $\h(t) \in \mathbb{R}^N$, where $t$ represents time.
The system is given by state-space equations~\cite{kalman1960new}
\begin{align}
	\frac{\D\h(t)}{\D t}
	&= A\h(t) + \B x(t)\label{steq}\\
	y(t)
	&= \C^\T\h(t)
\end{align}
where $A \in \mathbb{R}^{N \times N}$ is the state transition matrix, $\B$ and $\C$ have the same size of $\h$. These equations describe how the latent state evolves over time, determined by both the current state and the input.

The continuous-time state-space equations are discretized to approximate the system's behavior over a small time interval
$\Delta$~\cite{guefficiently,gu2023mamba}.
This process yields the discretized state-space equations
\begin{align}
	\h_l &= \bar{A}\h_{l-1} + \bar{\B}x_l \\
	y_l &= \C^\T\h_l
\end{align}
where $\bar{A}=e^{\Delta A}$, $\bar{\B}=(\Delta A)^{-1}(e^{\Delta A}-I)\Delta\B$, and $I$ is the identity matrix.

By iteratively applying the discretized state equations~\cite{guefficiently,gu2023mamba}, the output $y_L$ as a linear combination of the inputs is derived by,
\begin{align}
	y_L=[\C^\T A^{L}\bar\B,\ldots,\C^\T\bar A^2\bar\B,\C^\T A\bar\B]\x\label{convSSE}
\end{align}
where $\x=[x_1,x_{2},\ldots,x_{L}]^\T$ represents an input sequence of length $L$.

Then, we can obtain $\y=[y_1,y_{2},\ldots,y_{L}]^\T$ by
\begin{align}
	\y
	&=
	\left[
	\begin{array}{ccccc}
		\C^\T\oA\ob    &0                &\cdots & 0            &0  \\
		\C^\T\oA^2\ob   &\C^\T\oA\ob &\cdots & 0               &0     \\
		\vdots          &\vdots           &\ddots &\vdots   &\vdots \\
		\C^\T\oA^{^{L-1}}\ob&\C^\T\oA^{L-2}\ob   &\cdots      &\C^\T\oA\ob& 0\\
		\C^\T\oA^{L}\ob &\C^\T\oA^{L-1}\ob   &\cdots &
		\C^\T\oA^2\ob&\C^\T\oA\ob
	\end{array}
	\right]\x\,.
	\label{SSM}
\end{align}
Eq.~\eqref{SSM} can be efficiently accelerated using parallel block operations~\cite{gu2023mamba}, allowing it to achieve high processing speeds with reduced memory costs.

In \textit{vanilla} Mamba~\cite{gu2023mamba}, each element at position $l$ of the sequence $[\C^\T A^L\bar\B,\C^\T\bar A^{L-1}\bar\B,\ldots,\C^\T\bar A\bar\B]$ is designed to correspond to the $l$-th element $x_l$ of the input sequence. This correspondence achieved by expanding $\ob$, $\C$, and $\Delta$ along the sequence dimension. Additionally, both $\Delta$ and $\oA$ are extended to include the feature dimension of $D$, ensuring dimension consistency between input and output features. Then, due to dimensional adjustments, the notations of $\oA$, $\Delta$, $\ob$ and $\C$ are redefined by $\bm{A}\in\mathbb R^{D\times N}$, $\bm\Delta\in\mathbb R^{B\times L\times D}$, $\bm{B}\in\mathbb R^{B\times L\times N}$, and $\bm{C}\in\mathbb R^{B\times L\times N}$. The input and the output features, $\bm{F}_i$ and $\bm{F}_j$, both have dimensions of $\mathbb R^{B\times L\times D}$ where $\bm{F}_i$ and $\bm{F}_j$ are the inputs and outputs of the SSM. Here, $B$ denotes the mini-batch size, $D$ is the feature dimension, and $L$ is the sequence length. A selective SSM is defined by $\bm{F}_j={\rm SSM}(\bm{F}_i,\bm{A}_{\theta},\bm{B},\bm{C},\bm{\Delta}_{\theta})$, where $\bm{A}_{\theta}$ is defined as a learnable model parameter and $\bm\Delta_{\theta}$ serves as a learnable gated parameter.

\begin{figure}[!t]
	\centering
	\includegraphics[width=0.96\linewidth]{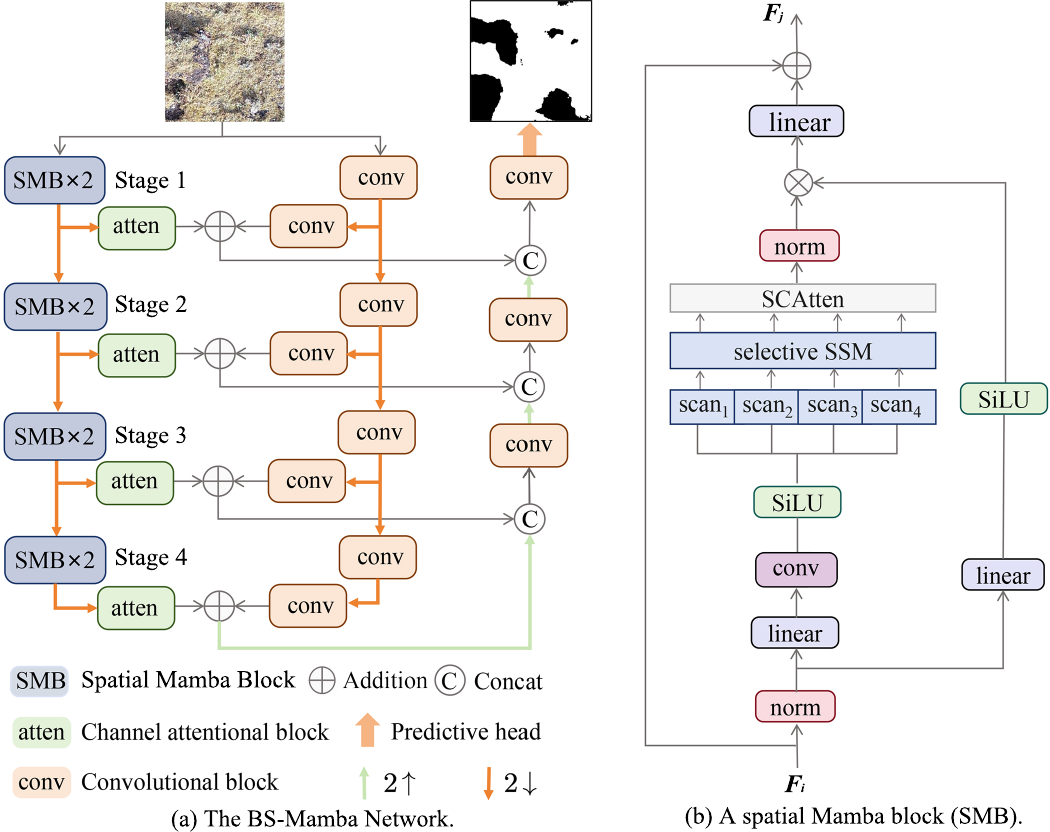}
	\caption{The BS-Mamba network utilizes a spatial Mamba block (SMB) to capture long-range dependencies and a convolutional block to retain spatial structure. The two-branch encoder provides the decoder with skip connections from each branch, enabling it to capture both long-range dependencies and fine-grained texture details in black-soil patches and the mattic epipedon.}
	\label{fig:BS-Mamba}
\end{figure}
\subsection{BS-Mamba}
Figure~\ref{fig:BS-Mamba}(a) provides an overview of the BS-Mamba network framework. High-resolution remote sensing images contain crucial spatial and long-range dependencies, which are vital for target detection tasks\cite{liu2024cascaded,zhu2024lrsnet,wang2022cctnet,he2022swin,10677460,wang2023bocknet,Wangcaai2023,cheng2024deep}. Since capturing long-range dependencies is essential for comprehensive feature extraction, we utilize spatial Mamba block (SMB), known for the ability to capture such dependencies effectively. Simultaneously, convolutional blocks, which good at preserving spatial structure, are incorporated to retain the fine-grained texture details of black-soil patches and the mattic epipedon. Specifically, the decoder leverages skip connections from the encoder  convolutional blocks to capture these textural features. To address these dual objectives effectively, we employ a two-branch encoder, allowing the decoder to receive skip connections from each branch. This design enables BS-Mamba to accurately capture both inter-patch relationships and texture details of black-soil patches and mattic epipedon.

To explore long-range dependencies, four-stage SMBs~\cite{huang2024localmamba} are employed in the first branch of the encoder. To make the best use of spatial structure, the second branch utilizes four-stage convolutional blocks. This combination allows the decoder to effectively explore long-range dependencies and retain the spatial texture details of black-soil patches and mattic epipedon, ensuring more accurate feature extraction while preserving essential spatial characteristics. We employ a hybrid encoder that integrates SMBs and convolutional blocks for efficient two-branch feature fusion. This hybrid approach enhances the BS-Mamba's robustness and adaptability to diverse black-soil patches. The decoder utilizes a cascaded upsampling for precise localization and incorporates skip connections to improve the detection accuracy of black-soil patches and the mattic epipedon. The convolutional blocks extract features at various resolutions, progressively refining image details. In parallel, the SMBs leverage SSM to capture global spatial dependencies through multidirectional scanning, complementing the convolution branch.

The outputs from the two branches undergo refinement through feature enhancement modules. Local features derived from the convolutional blocks are processed via an additional convolutional layer, which filters out redundant information. Concurrently, the channel attention module adjusts the global features extracted by the SMBs, selectively emphasizing or suppressing specific channels. These enhanced features are then fused through element-wise addition, effectively combining long-range dependencies with fine-grained texture details.

The BS-Mamba model processes input images, generating multiscale feature maps from both the convolutional and SMB branches. During the decoding phase, these feature maps are progressively upsampled and integrated using a cascade of upsampling, concatenation, and skip connections to restore spatial resolution. This refinement culminates in a final convolutional layer serving as the prediction head, accurately delineating the mattic epipedon regions.

Overall, the BS-Mamba model effectively captures fine-grained texture details and long-range dependencies across multiple scales. This design makes it particularly well-suited for detecting black-soil patches and the mattic epipedon in UAV imagery.
\subsubsection{Spatial Mamba Block}
Figure~\ref{fig:BS-Mamba}(b) shows an SMB of the BS-Mamba model. An image feature is divided into patches, which are flattened into vectors arranged these vectors sequentially to form $\bm F_i$. The SMB employs a novel local-scanning module~\cite{huang2024localmamba} to construct the sequence as input to the selective SSM. Initially, $\bm F_i$ is processed through layer normalization and passes through a linear layer, after which it splits into two parallel flows.

A depth-wise convolutional layer, similar to the one-dimensional convolution used in the \textit{vanilla} Mamba model, is then applied, followed by SiLU activation. The processed features are then directed into the local-scanning strategy~\cite{huang2024localmamba}, which captures spatial relationships and long-range dependencies through scans in four directions within local windows. This inter-patch scanning method maintains the structural integrity of 2D natural images while enhancing the model's ability to capture broad contextual relationships.

The SCAttn module~\cite{huang2024localmamba} uses spatial and channel attention to integrate the directional features into a unified output. SCAttn assigns importance across channel and spatial dimensions, emphasizing relevant details and filtering out redundant information, thereby enhancing the model's representational capacity.

At the end of an SMB, layer normalization and a gated scaling signal refine the output, ensuring coherent feature representation for downstream tasks.
\begin{figure}[!t]
	\centering
	\includegraphics[width=0.94\linewidth]{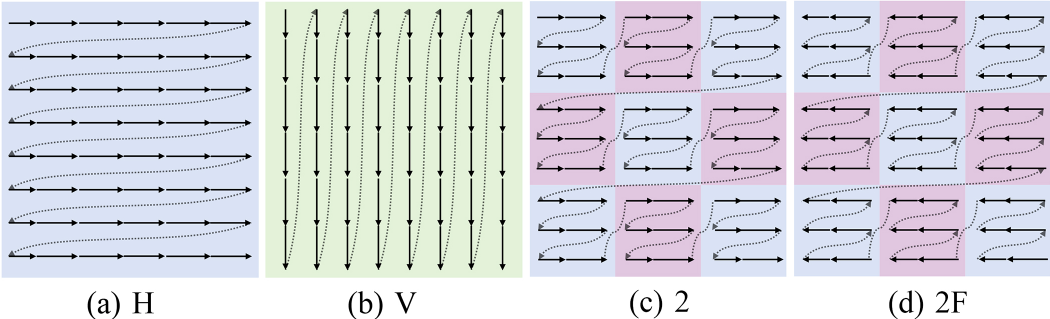}
	\caption{Four scanning strategies are illustrated:
		(a) `H' indicates horizontal scanning.
		(b) `V' represents vertical scanning.
		(c) `2' is scanning within a local $2 \times 2$ window.
		(d) `2F' is scanning in the flipped direction within a $2 \times 2$ window.  }
	\label{fig:LSM}
\end{figure}
\subsubsection{Local-Scanning Module}
We illustrate four different scanning strategies in Figure~\ref{fig:LSM}. The first strategy involves horizontal scanning, while the second adopts vertical scanning. The third utilizes horizontal scanning within a local $2\times2$ window, and the fourth applies flipped horizontal scanning within the same local $2\times2$ window. In this study, we primarily focus on local-scanning strategies.
The local-scanning strategies are integral to the SMB~\cite{huang2024localmamba}, enabling the effective extraction of global features from multiple directions within an image. These local scanning strategies not only captures long-range dependencies but also preserves the relationships between neighboring patches.
\begin{figure}[!t]
	\centering
	\includegraphics[width=0.94\linewidth]{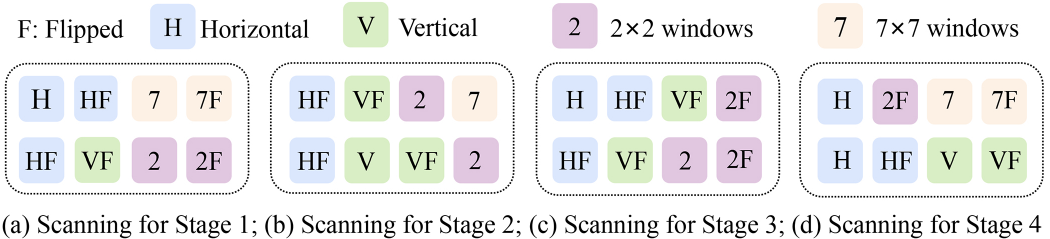}
	\caption{Scanning strategies in four stages of BS-Mamba: each stage contains two SMBs, with each SMB's SSM configured to scan sequence tokens in four distinct directions. In each stage, the first row represents the scanning configuration for the first SSM, while the second row corresponds to the second SSM in each stage.}
	\label{fig:LocalVim}
\end{figure}

To enhance BS-Mamba's capability in capturing long-range dependencies, we integrate some novel local-scanning strategies alongside the conventional horizontal and vertical scanning directions. Each SMB employs four scanning directions, with each stage containing two SMBs. This setup is illustrated in Figure~\ref{fig:LocalVim}, showing the four scanning directions for each SMB in every stage.
The local-scanning strategies group neighboring patches effectively, preserving the inherent spatial dependencies of the image~\cite{huang2024localmamba}. These patches are processed by the SSM, which captures both fine-grained spatial relationships and long-range dependencies within local windows.
This approach strikes a balance between capturing detailed local features and global contextual information, significantly enhancing the model's ability to extract large-scale spatial features from diverse orientations in UAV images. By balancing the relationships between neighboring patches and long-range dependencies, the method ensures robust feature extraction and improved global information representation.

\begin{algorithm}[!t]
	\caption{The algorithm for the BS-Mamba model.}
	\begin{algorithmic}[1]
		\Require Input tensor $\X$.
		\Ensure Output tensor $\Y$.
		\State \(\bm F_0 \gets \X\) 
		\For{$i = 1$ to $4$} \Comment{Mamba branch.}
		\State $\bm F \gets \text{SMB}(\text{SMB}(\bm F_{i-1}))$
		\State $\bm F_i\gets \text{pooling}(\bm F)$ \Comment{Downsampling.}
		\State $\bm F_i^{\rm ssm} \gets \text{atten}(\bm F_i)$ 
		\EndFor
		\State \(\bm F_0 \gets X\)
		\For{$i = 1$ to $4$} \Comment{Convolutional branch.}
		\State $\bm F \gets \text{conv}(\bm F_{i-1})$
		\State $\bm F_i \gets \text{pooling}(\bm F)$ \Comment{Downsampling.}
		\State $\bm F_i^{\rm conv} \gets \text{conv}(\bm F_i)$ 
		\State $\bm F_i^{{\rm fused}} \gets \bm F_i^{{\rm ssm}}+ \bm F_i^{{\rm conv}}$ \Comment{Feature fusion.}
		\EndFor
		\State $\bm F_4=\bm F_4^{\rm fused}$
		\For{$i = 3$ to $1$}\Comment{Decoder for reconstruction.}
		\State $\bm F_i \gets \text{conv}([\text{bilinear}(\bm F_{i+1}),\bm F_i^{\rm fused}])$
		\EndFor
		\State \( \bm Y \gets \text{conv}(\bm F_1) \)
	\end{algorithmic}
\end{algorithm}
\subsection{Overall Algorithm}
An algorithm for BS-Mamba model is presented in Algorithm 1. Initially, an input image $\X$ passes through the two-branch encoder. In the convolutional branch, the input image $\X$ is processed through a sequence of two consecutive $3\times3$ unpadded convolutions, each followed by a rectified linear unit (ReLU) activation and a $2\times2$ average pooling operation with a stride of $2$ for progressive downsampling. Four stages generates feature maps $\bm F_i,\forall\,i\in\{1,2,3,4\}$ with resolutions of $1$, $1/4$, $1/8$, and $1/16$ of the original resolution of $\X$.

Meanwhile, in the Mamba branch, the input image $\X$ is processed through four SMBs, as illustrated in Figure~\ref{fig:BS-Mamba}(b). Within this branch, $\X$ is flattened into a sequence of patch size  of four. This sequence is then refined through progressive downsampling by patch merging in Stage 1, Stage 2, Stage 3, and Stage 4, utilizing the local-scanning module and SSM to extract global spatial features across four directions, resulting in output feature maps $\bm F_i$ with resolutions of $1/4$, $1/8$, $1/16$, and $1/32$ of the original $\X$. To align with the corresponding stages in the convolutional branch, $\bm F_i$ is upsampled by a factor of four at each stage using linear interpolation. An SMB is detailed in \S2.3.1. To enhance the capacity of local features extracted from the convolutional blocks for more effective downstream processing, we employ convolutional layer to the output of each stage while maintaining the number of channels. To further enhance the expressive capacity of global features from the SMBs, we leverage a channel attention module, which involves using global average pooling followed by a linear layer to process $\bm F_i$. A subsequent ReLU activation function preserves non-linearity, and a sigmoid function generates channel weights, which are then multiplied with $\bm F_i$ to squeeze and excitation the channels of the feature maps. Through these two enhancement operations, the extracted features $\bm F_i^{{\rm ssm}}$ and $\bm F_i^{{\rm conv}}$ can represent  more effective and precise multiscale features.
Subsequently, we perform element-wise addition of $\bm F_i^{{\rm ssm}}$ and $\bm F_i^{{\rm conv}}$ along the channel dimension, enabling the fusion of two-branch features. For each stage, the fused features are obtained by $\bm F_i^{\rm fused}=\bm F_i^{\rm ssm}+\bm F_i^{\rm conv}$.

The decoder comprises a cascaded upsampling module and an additional convolutional block. The cascaded upsampling module comprises three stages, each incorporating bilinear interpolation for upsampling, feature concatenation, and two convolutional layers. The final convolutional block further refines features by performing a 2D convolutional operation to compress the channels.

In the first stage of decoder, $\bm F_4=\bm F_4^{\rm fused}$ is upsampled using bilinear interpolation to match the resolution of $\bm F_3$ while maintaining double the number of channels compared to $\bm F_3$. The upsampled $\bm F_4$ is concatenated with $\bm F_3^{\rm fused}$ along the channel dimension, producing a feature map $\bm F_3$ and twice the channel count of $\bm F_2$. The operation is represented as $\bm F_3 = {\rm conv}([{\rm bilinear}(\bm F_4), \bm F_3^{\rm fused}])$, where ${\rm conv}$ denotes a convolutional block composed of a $3\times3$ convolution, batch normalization, and ReLU. This process is repeated in the next stages, yielding $\bm F_2 = {\rm conv}([{\rm bilinear}(\bm F_3), \bm F_2^{\rm fused}])$ and $\bm F_1 = {\rm conv}([{\rm bilinear}(\bm F_2), \bm F_1^{\rm fused}])$, with $\bm F_1$ now at the original image resolution.Then, a convolutional layer and a softmax layer are applied to $\bm F_1$ to obtain the prediction $\bm Y$.

Finally, a channel-compression convolution layer is applied to produce a prediction map $\Y=[Y^{\rm blk},Y^{\rm mat}]$ with two channels. Since all of our dataset masks comprise two classes, we employ the cross-entropy loss~\cite{martinez2019taming}
\begin{align}
	\mathcal{L}_{\rm ce} = - \sum_{ij}  \left(
	t^{\rm blk}_{ij} \ln y^{\rm blk}_{ij} +
	t^{\rm mat}_{ij} \ln y^{\rm mat}_{ij}
	\right)\,.
\end{align}
For each channel, we compute the Intersection-over-Union (IoU) loss~\cite{zheng2020distance}:
\begin{align}
	\mathcal{L}_{\rm IoU} = 1 - \frac{\sum_{ij} t_{ij} y_{ij}}{\sum_{ij} t_{ij} + \sum_{ij} y_{ij} - \sum_{ij} t_{ij} y_{ij}}
\end{align}
where $T = [t_{ij}]$ denotes the ground truth of each channel. Since we have two channels, we use the mean IoU loss,
\begin{align}
	\mathcal{L}_{\rm mIoU} =
	\frac{\mathcal{L}_{\rm IoU}^{\rm blk} + \mathcal{L}_{\rm IoU}^{\rm mat}}{2}\,.
\end{align}
We combined the cross-entropy loss and mIoU loss in a weighted manner to ensure the model focuses on both classification accuracy and spatial consistency in segmentation.The overall loss function is given by
\begin{align}
	\mathcal{L} = \lambda_{1}\mathcal{L}_{\rm ce} + \lambda_{2}\mathcal{L}_{\rm mIoU}\,.\label{loss_total}\tag{10}
\end{align}
where $\lambda_{1}$ and $\lambda_{2}$ are two hyperparameters.
\section{Dataset Annotation}\label{sec:data}
\begin{figure}[!t]
	\centering
	\includegraphics[width=0.46\textwidth]{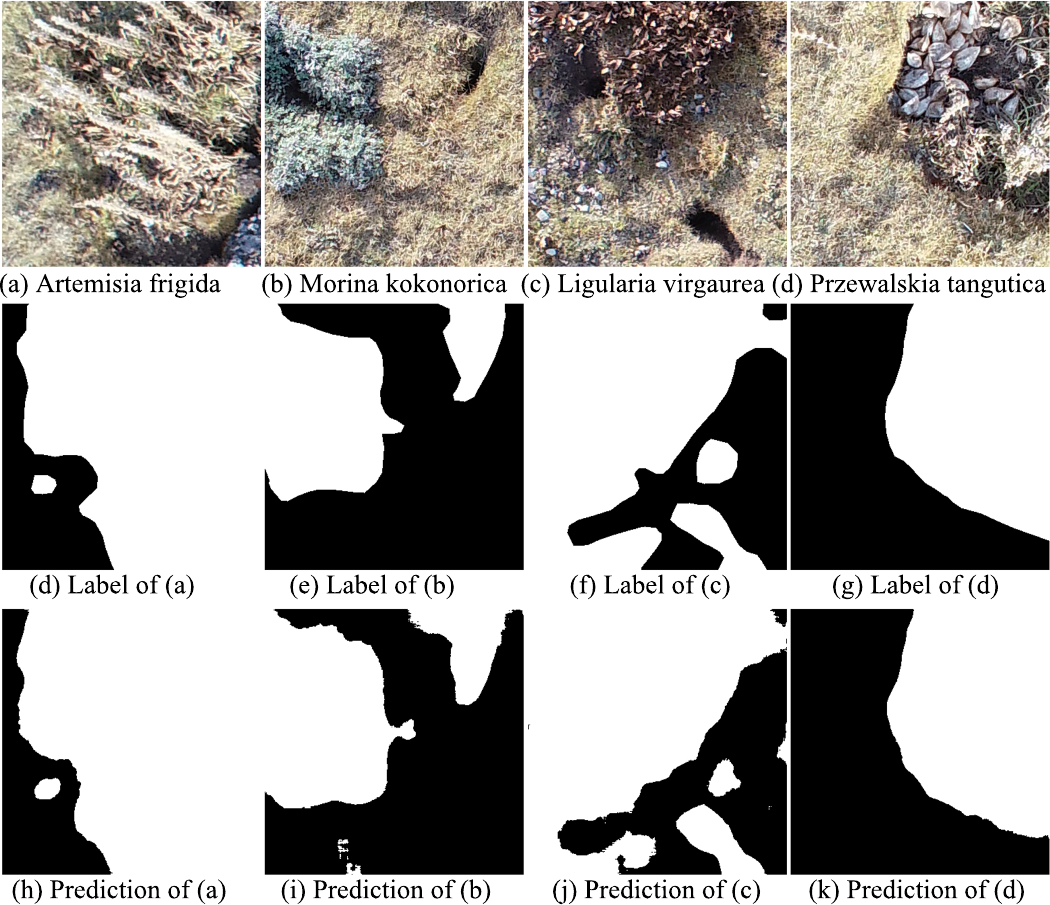}
	\caption{Different black-soil types of weeds and poisonous plants: The first row showcases original UAV-captured images highlighting diverse weeds and poisonous plants in black-soil areas.  The second row shows the ground truth labels. The third row presents the segmentation results generated by the BS-Mamba model, demonstrating its ability to accurately delineate these challenging features.}
	\label{fig:fig6}
\end{figure}

\begin{figure}[t]
	\centering
	\includegraphics[width=0.46\textwidth]{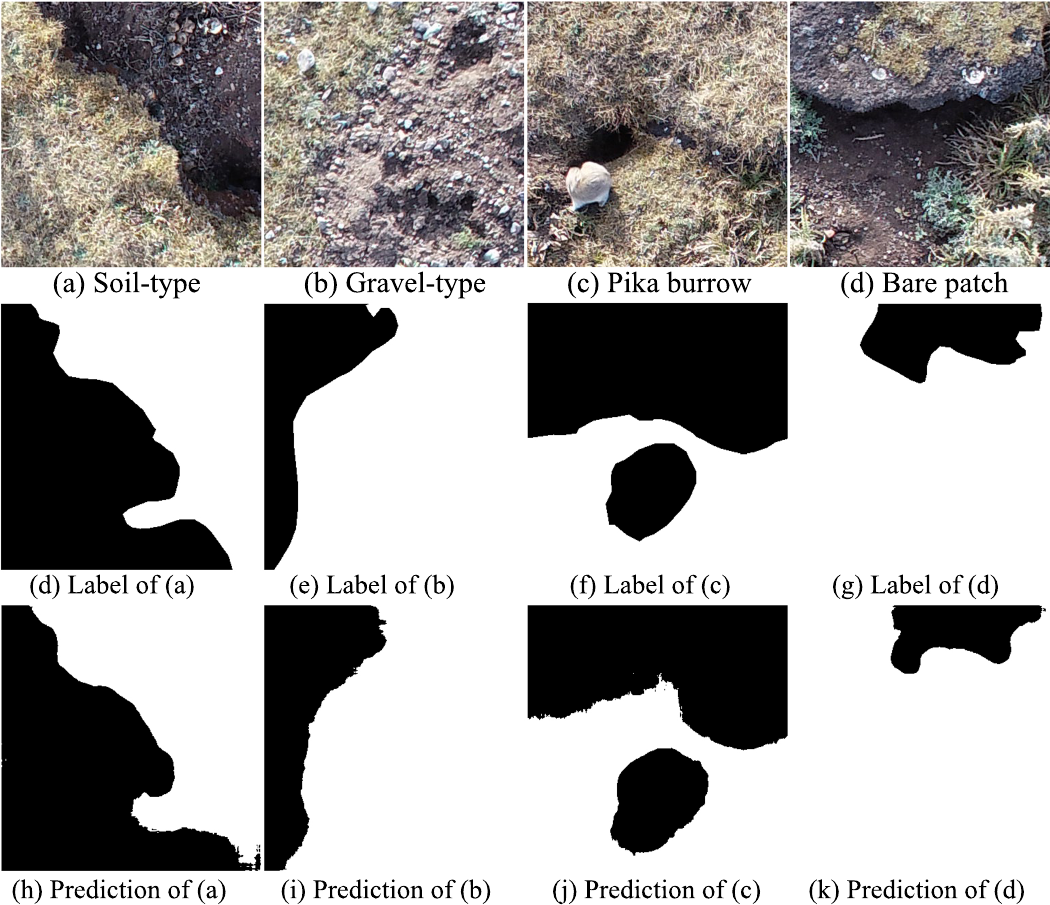}
	\caption{Different black-soil types: The first row displays raw UAV images of black-soil patches, capturing various soil and degradation types. The second row shows the ground truth labels. The third row presents the corresponding segmentation results generated by the BS-Mamba model.}
	\label{fig:fig5}
\end{figure}

In this paper, we introduce a well-annotated dataset tailored for black-soil area detection. The UAV images were acquired in Machin County, Guoluo Prefecture, Qinghai Province, China, at an average altitude of 3,588 meters, under harsh geographical conditions. These UAV-captured images encompass representative black-soil regions, including diverse soil, gravel, and rock types. They contain varying degradation stages, such as desertified grassland and severely desertified grassland~\cite{XBNX202302004}. Black-soil patches, commonly referred to as extremely degraded grasslands on the QTP where the mattic epipedon is absent, exhibit significant diversity. As a result, black-soil areas include various types based on degradation and landform characteristics. The annotated examples in our dataset reflect several typical types of black-soil areas. As shown in Figure~\ref{fig:fig5}, the dataset includes plateau pika burrows~\cite{GUO2012104} and barren patches~\cite{LIN-Hui-long_477}, alongside the ground-truth labels and BS-Mamba detection predictions. Furthermore, Figure~\ref{fig:fig6} highlights a variety of weeds and poisonous plants along with the ground-truth labels and BS-Mamba predictions~\cite{fan197,shang1157}. This dataset offers a resource for training, evaluating, and advancing models in black-soil area detection.

We name the provided dataset the QTP-BS dataset, focusing on black-soil area detection on the QTP. QTP-BS is meticulously designed to reflect the region's degraded landscapes, considering scenario diversity, vegetation density, and human activity impacts. Approximately 400 high-resolution images (\(5472 \times 3648\)) were captured from a real-world degraded area. These images represent a broad range of black-soil characteristics. Afterwards, we carefully selected 100 high-resolution images to ensure the dataset's representativeness of typical black-soil characteristics. Images with interference factors, such as shadows, blurriness, and other types of noise, were excluded during this selection. The chosen areas reflect the comprehensive characteristics of extremely degraded black-soil regions. These regions exhibit: (1) Uneven mattic epipedon distribution, with large mattic epipedon patches interspersed among exposed soil areas; and (2) Small snow-covered patches introducing additional complexity to the scenes.

The annotation process posed significant challenges, primarily due to the dataset's large scale and the inherent difficulty in distinguishing indistinct boundaries between extremely degraded black-soil patches and mattic epipedon. Under the guidance of ecological experts and after a thorough analysis of the black-soil patches, we applied meticulous techniques to finely label the images. Given the complexity of the scenes, annotation quality directly impacts the model's performance. To ensure accuracy, we undertook an iterative process involving repeated annotation, comparison, and cross-checking. This exhaustive effort spanned a full year, ensuring a dataset with manual pixel-level annotations of the highest precision. The resulting dataset combines large scale, broad coverage, and high accuracy, providing a robust foundation for advancing research and experimentation in black-soil area detection.
\section{Experimental Results}\label{sec:exp}
\subsection{Experimental Setting}
The high-resolution UAV remote sensing images in the QTP-BS dataset are referred to as scenes. These scenes are divided into smaller images of \(384 \times 384\) pixels. Each high-resolution scene produces 150 smaller images, resulting in a total of 15,000 images along with their corresponding masks. Based on a systematic analysis, the QTP-BS dataset is split as follows: 8,400 images (from 56 high-resolution scenes) are used for training, 3,600 images (from 24 high-resolution scenes) for validation, and 1,500 images (from 10 high-resolution scenes) for testing. Additionally, 10 high-resolution scenes (1,500 images) from a different region are annotated to serve as a secondary test set, aimed at evaluating the model's generalization capability. This splitting strategy ensures a comprehensive and fair assessment of various network models in black-soil detection tasks, enabling more accurate comparisons of their ability to detect degraded black-soil grasslands.

During training, the BS-Mamba backbone are initialized randomly. We perform optimization using the ADAM optimizer~\cite{kingma2014adam}  with an initial learning rate of 2e-4, a weight decay of 5e-4, and 60 epochs. Both, \(\lambda_1\) and \(\lambda_2\), are set to 0.5. We adopt the CutMix augmentation strategy~\cite{yun2019cutmixregularizationstrategytrain}, where regions are randomly cut and pasted among training images, with the labels mixed proportionally to the region. CutMix improves model robustness by enabling the network to learn from diverse feature sets~\cite{DBLP:journals/corr/HendrycksG16c,liang2017enhancing}.

To achieve the detection effect of real-world images and maintain data integrity, predicted images are reconstructed into high-resolution images. This step facilitates detection tasks on real-world data, ensures result consistency, and aids in directly comparing predictions with ground truth, while  allowing for qualitative observation of image quality.

For comprehensive evaluation, three key metrics are employed: accuracy (ACC), \(F_1\)-score (\(F_1\)), and intersection over union (IoU). Accuracy measures the proportion of correctly classified pixels across all classes, providing a broad performance overview, though it may be skewed by class imbalance~\cite{wang2023revisiting}. The \(F_1\)-score, the harmonic mean of precision and recall, offers a balanced metric by jointly considering false positives and false negatives. Finally, IoU quantifies the overlap between predicted and ground-truth pixels, providing a precise assessment of segmentation accuracy.
\subsection{Comparative Experiments}
To comprehensively evaluate the performance of BS-Mamba on the QTP-BS dataset, we conduct a series of comparative experiments against baseline and advanced segmentation models. The selected models include the classic U-Net~\cite{ronneberger2015u} and HRNet~\cite{sun2019deep}, which are widely used for image segmentation tasks, as well as recent models leveraging Mamba, such as Mamba-UNet~\cite{wang2024mamba}, LocalMamba~\cite{huang2024localmamba}, and UltraLight VM-UNet~\cite{wu2024ultralight}. To further assess the generalization capability of BS-Mamba, all models are also evaluated on the secondary test set derived from a distinct region.
\begin{table*}[!t]
	\centering
	\caption{Experimental results on the first test set. The bold values indicate the best result.}\label{tb1}
	\begin{tabular}{l|lll|lll|l}
		\toprule
		Method&IoU$^{\rm blk}$&IoU$^{\rm mat}$&mIoU&$F_1^{\rm blk}$&$F_1^{\rm mat}$&$\overline{F}_1$&ACC\\
		\midrule
		U-Net & 83.34 & 78.95 & 81.14 & 90.84 & 88.13 & 89.49 & 89.77\\
		HRNet & 83.15 & 78.39 & 80.77 & 90.74 & 87.79 & 89.27 & 89.58 \\
		LocalMamba & 80.74 & 76.39 & 78.57 & 89.27 & 86.50 & 87.89 & 88.18 \\
		Mamba-UNet & 83.07 & 78.36 & 80.71 & 90.68 & 87.77 & 89.22 & 89.54 \\
		UltraLight VM-UNet & 78.05 & 73.78 & 75.92 & 87.52 & 84.73 & 86.13 & 86.42 \\
		BS-Mamba (ours) & \textbf{84.61} & \textbf{79.91} & \textbf{82.26} & \textbf{91.61} & \textbf{88.75} & \textbf{90.18} & \textbf{90.50} \\
		\bottomrule
	\end{tabular}
\end{table*}

\begin{table*}[!t]
	\centering
	\caption{Experimental results on a secondary test set. The bold values indicate the best result.}\label{tb2}
	\begin{tabular}{l|lll|lll|l}
		\toprule
		Method&IoU$^{\rm blk}$&IoU$^{\rm mat}$&mIoU&$F_1^{\rm blk}$&$F_1^{\rm mat}$&$\overline{F}_1$&ACC\\
		\midrule
		U-Net & 62.14 & 75.89 & 69.02 & 75.98 & 85.85 & 80.92 & 82.44 \\
		HRNet & 63.83 & 75.51 & 69.67 & 77.51 & 85.67 & 81.59 & 82.73 \\
		LocalMamba & 61.01 & 75.24 & 68.12 & 75.29 & 85.45 & 80.37 & 81.95 \\
		Mamba-UNet & 61.41 & 75.45 & 68.43 & 75.66 & 85.60 & 80.63 & 82.17 \\
		UltraLight VM-UNet & 62.07 & 58.61 & 60.34 & 76.37 & 72.60 & 74.49 & 75.45 \\
		BS-Mamba (ours) & \textbf{66.81} & \textbf{76.62} & \textbf{71.72} & \textbf{79.81} & \textbf{86.46} & \textbf{83.14} & \textbf{84.04} \\
		\bottomrule
	\end{tabular}
\end{table*}
Tables~\ref{tb1} and~\ref{tb2} summarize the performance metrics obtained on the first QTP-BS test set and the secondary test set. In Tables~\ref{tb1} and~\ref{tb2}, superscripts (blk) and (mat) denote results specific to black-soil patches and mattic epipedon, respectively. BS-Mamba exhibits substantial improvements across all evaluation metrics, underscoring its capacity to effectively capture both long-range dependencies and intricate texture details within black-soil patches and mattic epipedons.

For a more intuitive comparison of black-soil detection performance across the evaluated networks, we generate predictive results for each model. These predictions are visually compared against the original images and their corresponding ground truth. Red markers highlight missed detections (i.e., areas of mattic epipedon present but not recognized), while green markers indicate false detections (i.e., non-mattic regions mistakenly identified as mattic epipedon). With its error correction mechanisms, BS-Mamba achieves a notably lower error rate and higher detection accuracy compared to other models.

\begin{figure}[!t]
	\centering
	\includegraphics[width=\linewidth]{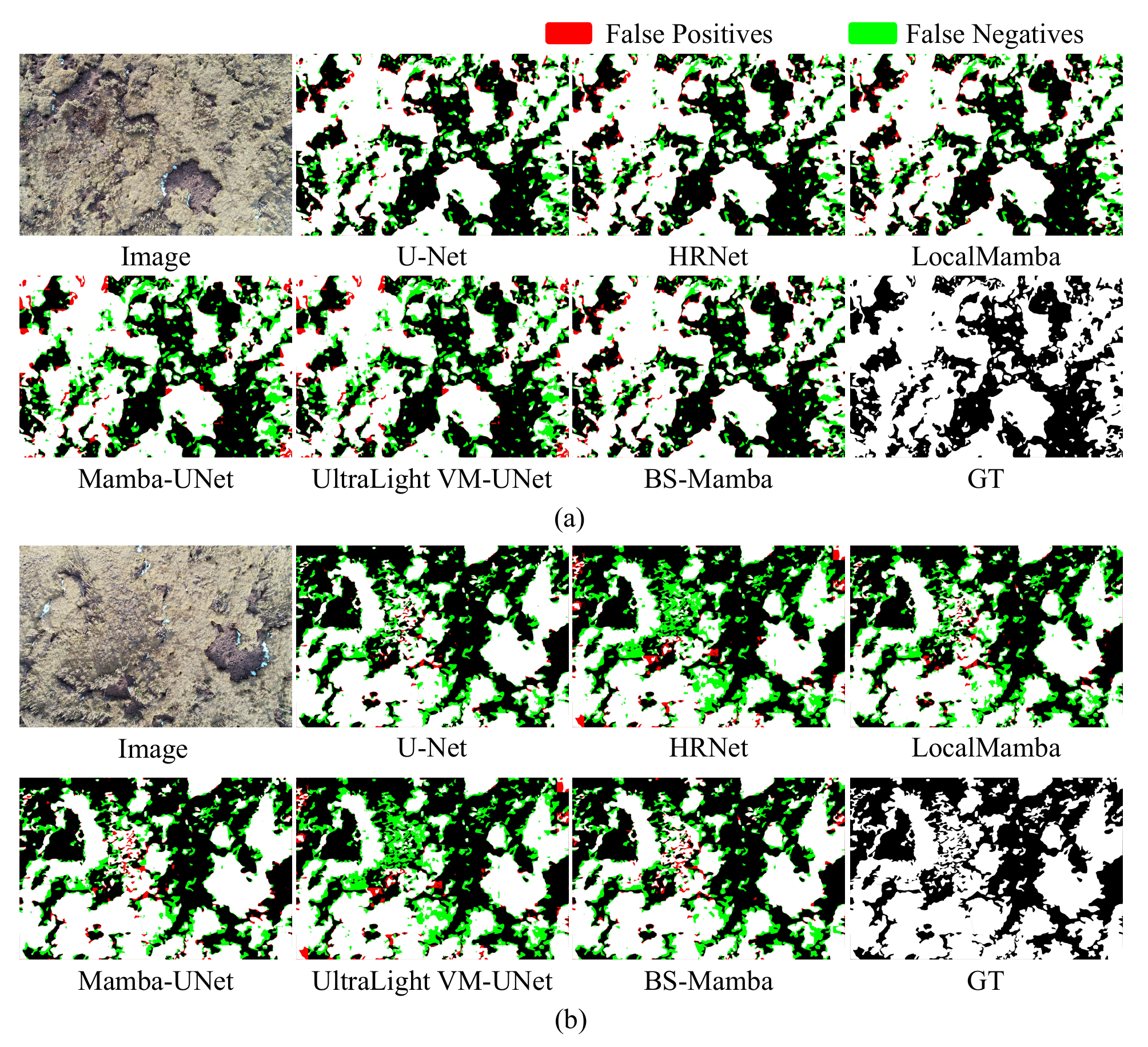}
	\caption{Visual comparison of BS-Mamba with baselines on black-soil area test set.}\label{fig:fig7}
\end{figure}

\begin{figure}[!t]
	\centering
	\includegraphics[width=\linewidth]{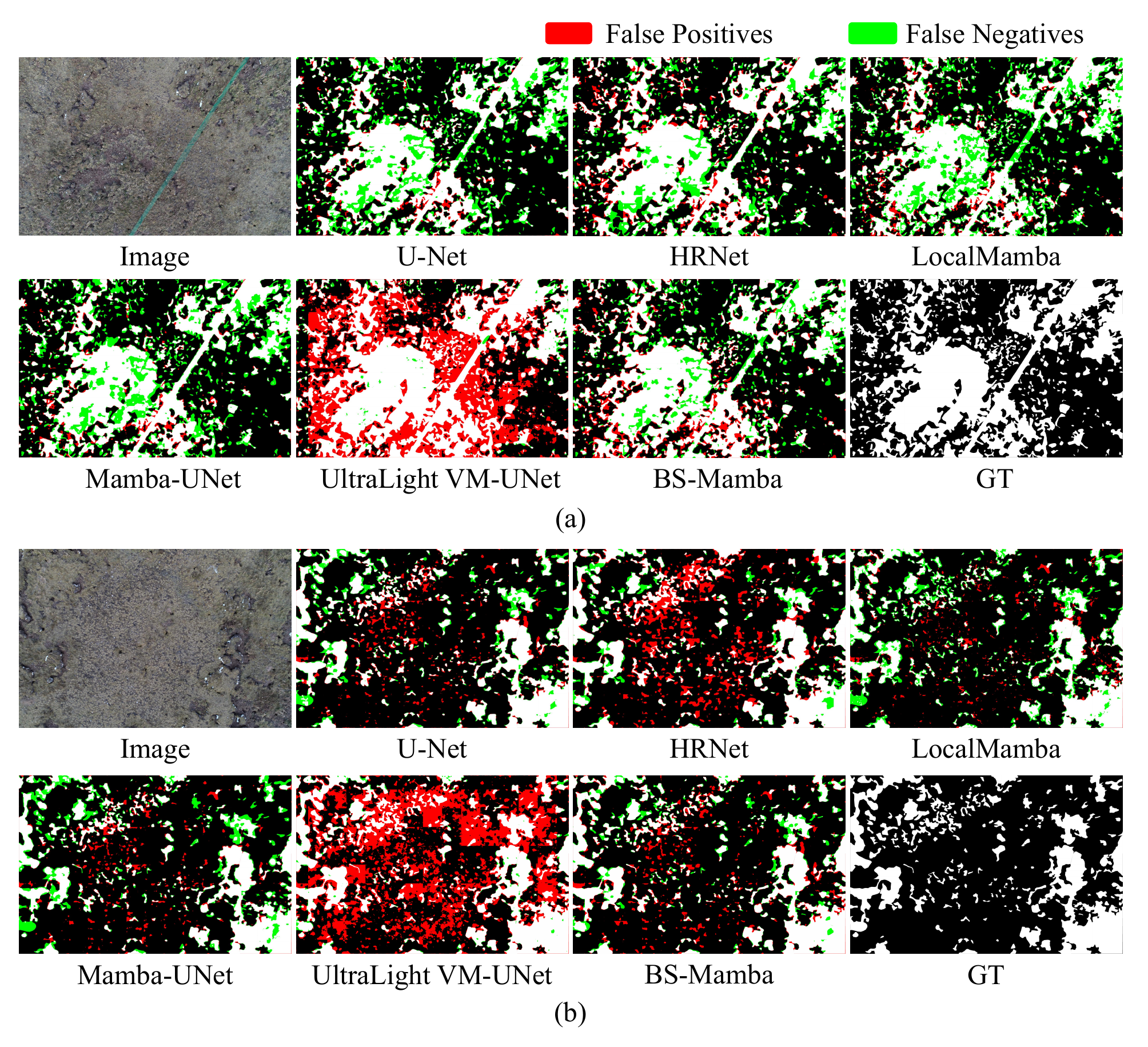}
	\caption{Visual comparison of BS-Mamba with baselines on black-soil area secondary test set.}
	\label{fig:fig8}
\end{figure}

We carefully select diverse real-world scenarios from two datasets for qualitative analysis, with each dataset providing two images, as shown in Figures~\ref{fig:fig7} and \ref{fig:fig8}. These images are derived from the primary region, representing transitional zones of extremely degraded grassland. These areas feature high vegetation cover, primarily composed of dwarf Stipa and small sedges, with an uneven distribution of species and moderately sized bare patches.

As illustrated in Figure~\ref{fig:fig7}(a), BS-Mamba's prediction masks exhibit significantly fewer false positives (red regions) and false negatives (green regions) compared to other models, such as U-Net and HRNet. This highlights its superior boundary delineation capabilities and enhanced detection accuracy. The model's outputs align closely with the ground truth, preserving intricate textures and achieving precise segmentation of fine-grained features. This effectiveness stems from the convolutional block, which maintains spatial integrity while leveraging skip connections to integrate detailed features from both the SMB and convolutional branches into the decoder. The combination of long-range dependencies with local spatial textures enables BS-Mamba to outperform other models in areas where balancing texture detail with global context poses challenges.
In Figure~\ref{fig:fig7}(b), while models like UltraLight VM-UNet exhibit a high density of false positives and false negatives, BS-Mamba maintains a more balanced performance. This balance is attributed to its convolutional block, which effectively preserves spatial structures, and its local-scanning strategy, which retains critical 2D dependencies while the SMB captures broader contexts. In homogenous soil regions, BS-Mamba produces cleaner outputs with fewer misclassifications than models such as Mamba-UNet and LocalMamba, demonstrating the strength of its two-branch encoder and decoder in robustly fusing local and global features while suppressing noise.

Figure~\ref{fig:fig8}(a) further validates BS-Mamba's superiority. It consistently outperforms U-Net and Mamba-UNet by generating significantly fewer false positives, especially in regions with intricate boundaries. Conversely, UltraLight VM-UNet introduces numerous false negatives, leading to under-segmentation. The SMB in BS-Mamba excels in capturing complex boundaries, particularly in areas with overlapping soil and vegetation. Its architecture, with two-branch encoding and skip connections, seamlessly integrates spatial and semantic features, enabling accurate identification of subtle patterns in heterogeneous landscapes, such as mixed soil and invasive vegetation, where other models struggle.

\begin{figure}[!t]
	\centering
	\includegraphics[width=0.94\linewidth]{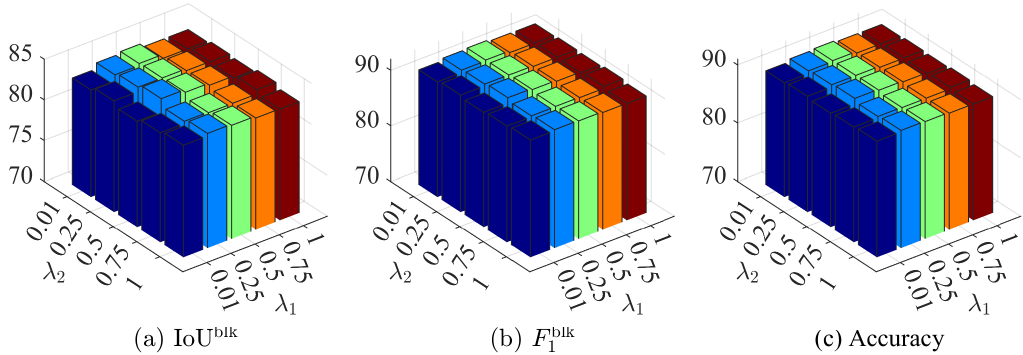}
	\caption{Sensitivity analysis of $\lambda_1$ and $\lambda_2$.}
	\label{fig:fig9}
\end{figure}

As demonstrated in Figure~\ref{fig:fig8}(b), BS-Mamba again outperforms competing models in regions with complex boundaries. By leveraging convolutional blocks and a channel attention mechanism, it refines spatial details and dynamically adjusts the relevance of feature channels. This capability is vital for differentiating between visually similar regions, such as mattic epipedon and black-soil patches. Despite being evaluated on an external dataset, BS-Mamba retains good performance, highlighting its adaptability to diverse scenarios and soil textures. The loss function of BS-Mamba, Eq.~\eqref{loss_total}, consists of two terms, each with its own trade-off parameter. Parameter insensitivity is crucial for enhancing the model's stability. The two parameters, $\lambda_1$ and $\lambda_2$, need to be determined. To illustrate the insensitivity to variations in $\lambda_1$ and $\lambda_2$, we conduct experiments evaluating IoU, \(F_1\), and accuracy as shown in Figure~\ref{fig:fig9}. Both parameters are varied within the range of $[0.01, 0.25, 0.5, 0.75,1]$. The results indicate stable performance across a wide range of parameter values, with the optimal results achieved when $\lambda_1$ and $\lambda_2$ are set to 0.5 and 0.25, respectively.

\subsection{Ablation Study}
To validate the effectiveness of various design choices, we conduct ablation studies.

\begin{figure}[!t]
	\centering
	\includegraphics[width=0.96\linewidth]{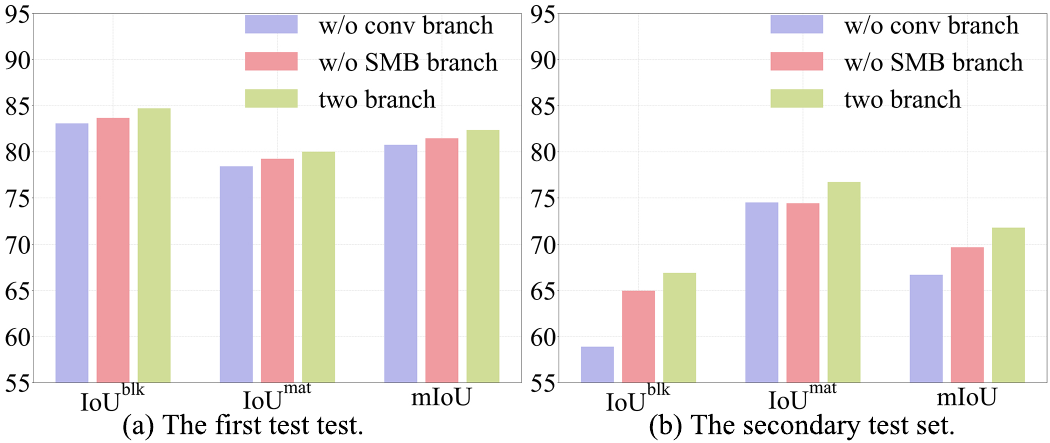}
	\caption{Ablation study of different types of branch setting on two test sets.}
	\label{fig:fig10}
\end{figure}

The first experiment focused on analyzing the feature extraction capabilities of different encoder configurations. Three experimental setups are evaluated: 1. SMBs as the sole encoder. 2. Convolutional blocks as the independent encoder. 3. The proposed two-branch encoder combining both SMBs and convolutional blocks. As illustrated in Figure~\ref{fig:fig10}, the results indicate that BS-Mamba, utilizing the proposed two-branch encoder, achieves the best performance by effectively balancing the strengths of both approaches. This hybrid encoder strategy enhances overall network performance. Convolutional branch outperform SMBs due to their superior ability to capture fine-grained texture details and preserve spatial structures. This finding highlights a key limitation of SMBs: treating images as 1D sequences prioritizes global context modeling but damages spatial structure, which cannot be fully recovered through direct upsampling.
\begin{figure}[!t]
	\centering
	\includegraphics[width=0.96\linewidth]{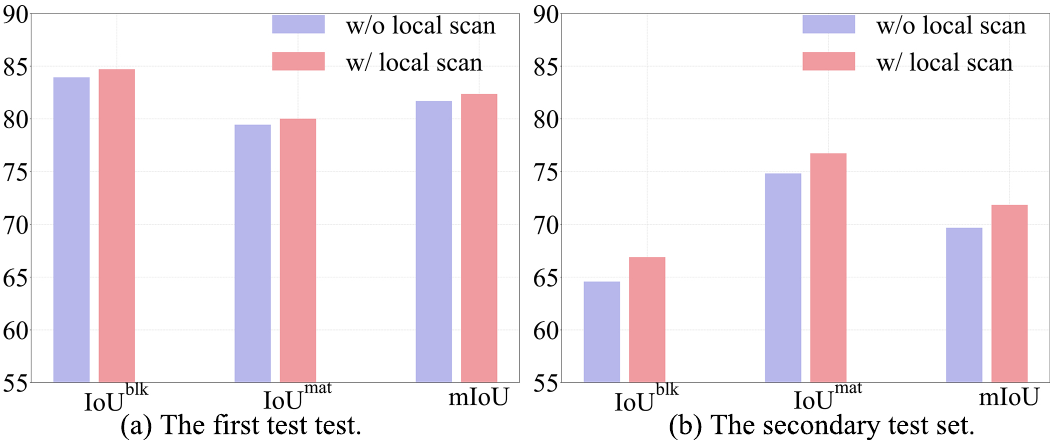}
	\caption{Ablation study of the local-scanning strategy on two test sets.}
	\label{fig:fig11}
\end{figure}

The second  ablation study examine the influence of the local scanning mechanism on segmentation. Specifically, we analyze the window scanning directions in the SMBs at each stage. As shown in Figure~\ref{fig:fig11}, the inclusion of a local-scanning module significantly improves segmentation results on both test sets. This module addresses a crucial limitation of treating non-causal image data as causal sequences, which disrupts relationships between neighboring patches. By employing local scanning, the information loss caused by traditional flattening is mitigated.
\section{Conclusion}\label{sec:con}
The paper presents a new approach to monitoring and managing the degradation of ``black-soil-type'' grasslands on the QTP, addressing an urgent ecological challenge using Mamba. To this end, we introduce the QTP-BS dataset, a large-scale, high-resolution dataset annotated by ecological experts. This dataset captures diverse degradation scenarios, providing a robust foundation for training and evaluating segmentation models.

We propose BS-Mamba, a novel model that combines the Mamba architecture's ability to model global dependencies with the spatial precision of convolutional operations. Its two-branch encoder effectively integrates long-range semantic features with fine-grained texture details, enabling the accurate delineation of black-soil patches and grass mattic epipedon. Experimental results on two independent test sets confirm that BS-Mamba significantly outperforms state-of-the-art models.
\section*{Disclosures}
The authors declare that there are no financial interests, commercial affiliations, or other potential conflicts of interest that could have influenced the objectivity of this research or the writing of this paper.
\section*{Code, Data, and Materials Availability}
The source code is available at \url{https://github.com/kunzhan/BS-Mamba}. The download link for the QTP-BS dataset is provided in README.md of the GitHub repository.
\section*{Acknowledgments}
This work was supported by Natural Science Foundation of Qinghai Province of China under No.~2022-ZJ-929.
{
\small
\bibliographystyle{ieeenat_fullname}
\bibliography{cvpr2025}
}



\end{document}